\input texinfo @c -*-texinfo-*- $Id: memt.texi,v 2.17 1996/12/31 13:00:59 ristad Exp $
@setfilename memt.info
@settitle Maximum Entropy Modeling Toolkit
@set VERSION 1.3
@set RELEASE 1.3 Beta
@set UPDATED December 31, 1996
@setchapternewpage on
@footnotestyle end
@syncodeindex vr cp
@syncodeindex fn cp
@syncodeindex tp cp

@titlepage
@sp 10
@title{Maximum Entropy Modeling Toolkit}
@subtitle Draft @value{UPDATED}
@subtitle Version @value{RELEASE}
@author Eric Sven Ristad <ristad@@cs.princeton.edu>
@page
@center{Copyright Notice, License and Disclaimer}

Permission to use, copy, and distribute this software and its
documentation without fee for not-for-profit academic or research
purposes is hereby granted, provided that the above copyright notice
appears in all copies, and that both the copyright notice and this
permission notice and warranty disclaimer appear in supporting
documentation, and that use of this software for research purposes is
explicitly acknowledged and cited in all relevant reports and
publications via the following citation form

@example
  Eric Sven Ristad. 
  Maximum entropy modeling toolkit, release @value{RELEASE}. 
  ftp://ftp.cs.princeton.edu/pub/packages/memt
  @value{UPDATED}.
@end example

The author disclaims all warranties with regard to this software,
including all implied warranties of merchantability and fitness.  In
no event shall the author or any entities or institutions to which
he/she is related be liable for any special, indirect or consequential
damages or any damages whatsoever resulting from loss of use, data or
profits, whether in an action of contract, negligence or other
tortious action, arising out of or in connection with the use or
performance of this software.

@vskip 0pt plus 1filll
Copyright @copyright{} 1996 Eric Sven Ristad. 
@end titlepage

@ifinfo
@node Top,  (dir),  (dir),  (dir)
@top Maximum Entropy Modeling Toolkit
The Maximum Entropy Modeling Toolkit supports parameter estimation and
prediction for conditional models over discrete domains in the maximum
entropy framework.  
@cindex version
This is version @value{RELEASE}, last updated @value{UPDATED}.

@menu
* License::             License, Copyright Notice, and Disclaimer
* Overview::            Overview
* Users Guide::         Users Guide
* Executables::         Executables
* File Formats::        File formats
* Background::          Background for maximum entropy modeling
* Index::               Function and concept index
@end menu

Copyright @copyright{} 1996 by Eric Sven Ristad.  All Rights Reserved.

@node License,  Overview,  ,    Top
@chapter License, Copyright Notice, and Disclaimer

@cindex copyright
Copyright @copyright{} 1996 by Eric Sven Ristad.  All Rights Reserved.

@cindex license
Permission to use, copy, and distribute this software and its
documentation without fee for not-for-profit academic or research
purposes is hereby granted, provided that the above copyright notice
appears in all copies, and that both the copyright notice and this
permission notice and warranty disclaimer appear in supporting
documentation, and that use of this software is explicitly
acknowledged and cited in all relevant reports and publications via
the following citation form:
@cindex citation
@example
  Eric Sven Ristad. 
  Maximum entropy modeling toolkit, release @value{RELEASE}. 
  ftp://ftp.cs.princeton.edu/pub/packages/memt
  @value{UPDATED}.
@end example

@cindex disclaimer
The author disclaims all warranties with regard to this software,
including all implied warranties of merchantability and fitness.  In
no event shall the author or any entities or institutions to which
he/she is related be liable for any special, indirect or consequential
damages or any damages whatsoever resulting from loss of use, data or
profits, whether in an action of contract, negligence or other
tortious action, arising out of or in connection with the use or
performance of this software.

@end ifinfo

@node   Overview,   Users Guide,    License,    Top
@chapter Overview

The Maximum Entropy Modeling Toolkit supports parameter estimation and
prediction for statistical language models in the maximum entropy
framework.  The maximum entropy framework provides a constructive
method for obtaining the unique conditional distribution p*(y|x) that
satisfies a set of linear constraints and maximizes the conditional
entropy H(p|f) with respect to the empirical distribution
f(x).@footnote{The conditional entropy H(p|f) = sum_<x,y> f(x)p(y|x)
log p(y|x) is taken with respect to the empirical distribution f(x).}
The maximum entropy distribution p*(y|x) also has a unique parametric
representation in the class of exponential models, as m(y|x) = r(y|x)
/ Z(x) where the numerator r(y|x) is a product of exponential weights
@example
        r(y|x) = prod_i alpha_i^g_i(x,y) 
@end example
for alpha_i = exp(lambda_i), and the denominator Z(x) = sum_y r(y|x) is
required to satisfy the axioms of probability.

This manual explains how to build maximum entropy models for discrete
domains with the Maximum Entropy Modeling Toolkit (MEMT).  First we
summarize the steps necessary to implement a language model using the
toolkit.  Next we discuss the executables provided by the toolkit and
explain the file formats required by the toolkit.  Finally, we review
the maximum entropy framework and apply it to the problem of
statistical language modeling.

@menu
* Availability::
* Mailing Lists::
* Acknowledgments::
@end menu

@node  Availability, Mailing Lists, , Overview
@section Availability

This is version @value{RELEASE}, last updated @value{UPDATED}.  
It is currently available from:
@example
        ftp://ftp.cs.princeton.edu/pub/packages/memt/
@end example
This release of the MEMT includes 
documentation in the following formats
@example
        ./doc/memt.@{info,html,ps@}
@end example
and binaries for the following Unix architectures:
@example
        ./bin/alpha             DEC Alpha OSF/1 3.0 (aka DEC Unix)
        ./bin/hppa              HP PA-RISC 9000 hpux 9.05
        ./bin/sgi               SGI MIPS IRIX 5.3
        ./bin/sun5              Sun SPARC SunOS 5.5
        ./bin/sun4              Sun SPARC SunOS 4.1.3
@end example

@node  Mailing Lists, Acknowledgments, Availability, Overview
@section Mailing Lists

@cindex announcements
If you would like to hear about new releases of the MEMT, then subscribe
to the memt-announce mailing list by sending mail to
majordomo@@cs.princeton.edu with the single line:
@example
        subscribe memt-announce
@end example

@cindex help
If you are encountering difficulty using the MEMT, or would like to
discuss maximum entropy modeling issues with skilled users, then send
mail to memt-help@@cs.princeton.edu.  If you are a skilled user of the
MEMT, or knowledgeable about maximum entropy modeling, then we encourage
you to subscribe to this majordomo mailing list and help field the
questions that are posted.

@node  Acknowledgments, , Mailing Lists, Overview
@section Acknowledgments

The author is grateful to Sanjeev Khudanpur, Harry Printz, and Salim
Roukos for helpful discussions about maximum entropy modeling.  A draft
version of the toolkit was created at the 1996 DoD Speech Recognition
Workshop at Johns Hopkins University.  The author is partially supported
by NSF Young Investigator Award IRI-0258517.

The toolkit was implemented using the Library of Practical Abstractions
(Ristad and Yianilos, 1996).  Documentation was created in GNU texinfo
format, from which info and postscript formats were derived directly;
html format was derived using texi2html from Lionel Cons.

@node  Users Guide,    Executables,   Overview,   Top
@chapter Users Guide

Only three simple steps are required to obtain a maximum entropy
distribution for a discrete conditional space Y|X -- define your
features, calculate their target expectations, and then find a model
m*(y|x) for the maximum entropy distribution.  Here we briefly guide you
through these three steps in the Maximum Entropy Modeling Toolkit.

The first step is to define a set G of features on Y|X.  Each n-ary
feature g_i(x,y) partitions the Y|X space into n equivalence classes,
where the j^th class consists of all events y|x where g_i(x,y) = j.
Choosing the features G defines the class R
of all exponential models over the features G, 
@example
        R = @{ m: m(y|x) = r(y|x) / Z(x) @} 
@end example
where 
@example
        r(y|x) = prod_i alpha_i^g_i(x,y)
        Z(x) = sum_y r(y|x).
@end example
and each alpha_i = exp(lambda_i).

The second step is to calculate the vector a = a_1 ... a_k of target
feature expectations from the training corpus.  The simplest and most
effective approach here is to set the target expectations to be the
empirical expectations.  The empirical expectation f[g_i] is the
expectation of the feature g_i(x,y) with respect to the empirical
distribution f(x,y) defined by the training corpus.@footnote{Our
notation p[g(x,y)] for the expectation of a function g(x,y) with respect
to a distribution p(x,y) departs from traditional notation E_p[g(x,y)].}
@example
        f[g_i] = sum_<x,y> f(x,y) g_i(x,y) 
@end example
Together, the features G and their target expectations define the class P
@example
        P = @{ p: p[g_i] = a_i for all i = 1...k @} 
@end example
of all probability functions that are consistent with training corpus as
viewed through our features G.  Recall that the intersection of P and R
includes only one model, the model for the maximum entropy distribution
p*(y|x) in P.  Therefore, the act of choosing the target expectations
also indirectly specifies our desired maximum entropy distribution
p*(y|x).

The third step is to perform maximum likelihood estimation on the
exponential model class R.  This identifies the maximum likelihood
exponential model m*(y|x) in R, which is equivalent to the desired
maximum entropy distribution p*(y|x) in P.  

Let us consider how to take these three simple steps in the MEMT.
@menu
* Estimation::          How to estimate model parameters
* Prediction::          How to evaluate testing data
* Common Errors::       Common errors and how to avoid them
@end menu

@node Estimation, Prediction, , Users Guide
@section Parameter Estimation

The first step is to specify the input/output behavior of your features.
This is accomplished with an events file.  An events file consists of
two parts, one for marginal events y in Y and the other for conditional
events y|x in Y|X.  For each marginal event y, the events file lists all
features that are active for y, independent of the context.  These are
the marginal features.  The events file must also list all the remaining
(ie., non-marginal) features that are active for every conditional event
y|x.  There are a great many conditional events and it would not be
feasible to list them all.  Fortunately, we need only list the
conditional events y|x that occur with positive frequency in the
training corpus.  This allows the toolkit to calculate the likelihood of
the training data according to a given model, and to calculate the
empirical distribution f(x) over the contexts X.  (Recall that the
empirical distribution f(x) is necessary to calculate the expectation of
each feature with respect to the joint model p(x,y) = f(x)p(y|x).)  We
also need to list all the events y|x that trigger a conditional feature,
but only for those contexts x that occur with positive frequency in the
training corpus.  This information is necessary to calculate the feature
expectations during parameter estimation.

The second step is to set your target expectations.  This is
accomplished with a parameters file.  As its name suggests, the
parameters file also contains the parameter value associated with each
feature.  An events file and an accompanying parameters file are all
that you need to start searching for the maximum entropy distribution
p*(y|x) in P.

Once you have constructed an events file and a parameters file, first
run the @code{me.memory} executable to make sure that your computer has
enough memory to use the toolkit.  You will need approximately as much
memory as the size of your events file to verify and train your model.
Next, run the @code{me.checker} executable to verify the correctness of
your files.  We recommend using the @var{-v} `verbose' option.  Don't
ignore the warnings!  When you are confident in the correctness of your
parameters and events files, then you are ready to take the third step.
Use the @code{me.estimate} executable to find the maximum likelihood
model m*(y|x) in the class R of exponential models over the features G.
Recall that the maximum likelihood exponential model m*(y|x) is
equivalent to the maximum entropy distribution p*(y|x) in P.  We
recommend using the @var{-m} `monotonic' option.  @code{me.estimate}
implements the highly effective improved iterative scaling algorithm
(Della Pietra et.al., 1995).

@node Prediction, Common Errors, Estimation, Users Guide
@section Prediction

Once you have found a model for the maximum entropy distribution, you
will want to evaluate the probabilities of future events according to
your model.  The simplest models will only require us to multiply a
series of conditional probabilities.  More sophisticated models, such as
those with hidden variables, may also require us to sum the
probabilities of complex events.  In order to do this with the toolkit,
you must create an expressions file that describes your desired
probability computation.  The operations employed in this probability
computation are (i) to evaluate the conditional probability p(y|x) of an
event, (ii) to multiply the results of other probability computations,
and (iii) to add the results of other probability computations.  You
must also create an events file for each expressions file.  This events
file contains all marginal events y that activate a marginal feature, as
well as all conditional events y|x that activate a conditional feature
and whose context x occurs in the expressions file.

Having constructed an expressions file and the corresponding events file
for your testing data, you are ready to evaluate the probability
expressions in your expressions file using the maximum entropy
distribution p*(y|x) obtained from your training data.  First use the
@code{me.checker} executable to verify the correctness of your new
expressions and events files as well as your trained parameters file.
Then you will use the @code{me.evaluate} executable to evaluate your
probability expressions.

@node Common Errors, , Prediction, Users Guide
@section Common Errors

@cindex debugging
It is quite easy to create an invalid or inconsistent model in the MEMT.
One way to create an invalid model is to use an inconsistent equivalence
classification for either the histories or the futures.  A second way to
create an invalid model is to provide a parameters file with incorrect
or inconsistent target expectations.  Let us briefly warn against both
errors.

@menu
* Equivalence Relations::
* Target Expectations::
@end menu

@node Equivalence Relations, Target Expectations, , Common Errors
@subsection Equivalence Relations

It is important not to confuse the equivalence relations used to define
X and Y with those used to define the behavior of your features.

The MEMT requires a denumerable domain X and a finite domain Y.  The
larger these domains are, the more statistically and computationally
costly the resulting model will be.  Therefore, the user should strive
to reduce the number of contexts X and symbols Y as much as possible.
The user must be careful, however, that every history is mapped to a
unique context x in X, and every future is mapped to a unique symbol y
in Y.  If two distinct histories h and h' are mapped to the same context
x, then the user must be sure that no feature can distinguish h and h'.
Similarly, if two distinct futures d and d' are mapped to the same
symbol y, then no feature can distinguish d and d'.

It is perfectly acceptable for distinct @emph{features} to overlap, that
is, for many features to be activated by a single conditional event y|x.
It is not, however, acceptable for any feature to take different values
for the same conditional event y|x on different occasions.

@node Target Expectations, , Equivalence Relations, Common Errors
@subsection Target Expectations

In our experience, the easiest way to stray from the path of effective
maximum entropy modeling is to @emph{not} use the empirical expectations
for your targets.  Therefore, we recommend that your target expectations
always be the strict empirical expectations.  The surest sign of an
invalid or inconsistent model is that it fails to converge during
parameter estimation.
@cindex convergence
Accordingly, the @code{me.estimate} executable provides four diagnostics
for convergence: d(m[g_i],a_i), |Update|, Max(alpha), and L(C|m).  Each
of these diagnostics has a straightforward interpretation.

@cindex d(m[g_i],a_i)
The first diagnostic d(m[g_i],a_i) is the Euclidean distance between the
feature expectations m[g_i] with respect to the current model m(x,y) =
m(y|x)f(x) and their target expectations a_i.  
@cindex |Update|
The second diagnostic |Update| is the magnitude of the lambda parameter
update vector.  Both of these quantities should be monotonically
decreasing with each iteration of @code{me.estimate}, even if the target
expectations are not the empirical expectations or if the targets are
mildly inconsistent.  If either of these quantities do not approach
zero, then we recommend you choose more conservative target
expectations.

@cindex Max(Alpha)
The third diagnostic Max(alpha) is the value of the maximal parameter
alpha_i = exp(lambda_i).  This value may fluctuate, but it should not
grow unreasonably large or small.  If Max(alpha) approaches zero or
grows without bound to +Inf, then your target expectations are
dangerously inconsistent and you would be better served to choose more
conservative targets.

@cindex L(C|m)
The fourth diagnostic L(C|m) is the total codelength of the corpus C
according to the current model m, that is, the negative log likelihood
of the corpus according to the current model.  This quantity should be
monotonically decreasing if the target expectations are the empirical
expectations.  If, however, your targets are smoothed or inconsistent,
then L(C|m) will not be monotonically decreasing.  If L(C|m) fluctuates
wildly for your model, then your target expectations are inconsistent,
and you should choose more conservative targets.  If your targets are
carefully smoothed empiricals, then L(C|m) may decrease for several
iterations, and then gradually increase for the remaining iterations.
In such a situation, you may wish to use the @var{-m} `monotonic' option
for @code{me.estimate}, to terminate estimation when L(C|m) stops
decreasing.  In our experience, the best model performance is obtained
when the codelength is minimized.

All of these potential problems may be avoided by using the strict
empirical expectations for your targets.  In our experience, using
empirical targets also provides the best model performance.

@node   Executables,    File Formats,   Users Guide,   Top
@chapter Executables 

The MEMT includes the following executables.

@findex me.memory
@deffn Executable me.memory
Prints the total available process memory on @var{stdout}.
@end deffn

@findex me.checker
@deffn Executable me.checker [-v] [-p @var{model}] [-e @var{events}] [-x @var{expressions}]
Given a parameters file @var{model}, an events file @var{events}, and/or
an expressions file @var{expressions}, @code{me.checker} verifies that
the given file(s) satisfy all syntactic requirements.  Errors and
warnings are written to @var{stderr} and a summary of the verification
process is written to @var{stdout}.  If the @var{-v} "verbose" option is
present, then more detailed messages are used.  @code{me.checker}
returns success (0) if and only if all files are compatible.
@end deffn

@findex me.estimate
@deffn Executable me.estimate [-m] @var{model.in} @var{events} @var{n} @var{model.out} 
Given a parameters file @var{model.in} and an events file @var{events},
@code{me.estimate} performs @var{n} iterations of improved iterative
scaling and writes the revised parameters to @var{model.out}.  If the
@var{-m} "be monotonic" option is present, then @code{me.estimate}
terminates when the codelength L(C|m) of the corpus increases.  Note
that premature termination can only happen if the target expectations in
the @var{model.in} parameters file are inconsistent.  @code{me.estimate}
also writes the following convergence information to @var{stdout} on
each iteration.  d(m[g_i],a_i) is the Euclidean distance between the
vector of model feature expectations m[g_i] and the vector of target
feature expectations a_i.  |Update| is the magnitude of the lambda
parameter update vector.  Max(alpha) is the value of the maximal alpha
parameter.  L(C|m) is the codelength of the corpus according to the
current model.  H(m|f) is the conditional entropy of the current model
m(y|x) with respect to the empirical distribution f(x).@footnote{Since
the H(m|f) diagnostic is costly to compute, it is only available in the
safe version of the @code{me.estimate} executable.}  Errors and warnings
are written to @var{stderr}.
@end deffn

@findex me.evaluate
@deffn Executable me.evaluate @var{model} @var{events} @var{expressions} @var{results}
Given a parameters file @var{model}, an events file @var{events}, and an
expressions file @var{expressions}, @code{me.evaluate} writes the
outcome of the computation specified in @var{expressions} to
@var{results} in nats, that is, as negative log likelihoods in base e.
The events files must include all marginal events y where some marginal
feature g_j(y) is nonzero and all conditional events y|x where some
conditional feature g_i(x,y) is nonzero and the context x occurs in the
expressions file.  Errors and warnings are written to @var{stderr}, and
status information is written to @var{stdout}.
@end deffn

@cindex safe
@cindex unsafe
@cindex debugging
@cindex errors
With the exception of the diagnostics @code{me.memory} and
@code{me.checker}, all executables are available in safe and unsafe
versions.  The safe versions include a significant amount of error
checking, full symbol tables for debugging, and no compiler
optimizations.  The unsafe versions are compiled with full compiler
optimizations, no symbol tables, and all error checking removed.  The
unsafe versions may also display less diagnostic information.  We
strongly recommend that you use the safe versions until you are
confident in the correctness of your files.  If you are using an unsafe
executable and it terminates prematurely, then you may need to run the
safe version to determine which error occurred.

@cindex core
@cindex memory
If one of the MEMT executables dumps core, then it is likely that your
machine does not have enough memory for your events file.  The amount of
available memory depends on the amount of real memory (ram) and virtual
memory (swap) available on the machine as well as on the kernel limits,
your own shell limits, and the memory usage of other active processes.
@findex unlimit
(At the very least, you should type @code{unlimit} in your shell and try
again.)
@vindex datasize
If your kernel @var{datasize} limit is too low, then you may need to
recompile your kernel.  The cheapest way to increase the amount of
available memory is to add more swap space, either by creating a larger
swap file or by adding another swap partition.

@node   File Formats,   Background,     Executables,    Top
@chapter File Formats

The maximum entropy toolkit specifies three ASCII file formats.  A
parameters file stores model parameters and the target expectations of
features.  An events file defines the behavior of the features on the
observed events.  An expressions file contains a series of probability
expressions, each of which evaluates to a probability value.  A
parameters file and an events file for the training corpus are necessary
to find a model for the desired maximum entropy distribution.  A
parameters file, an expressions file, and an events file for the testing
corpus are necessary to make predictions with the maximum entropy
distribution.

@tindex int
@tindex long
@tindex float
@tindex double
All @var{float} and @var{double} values are ASCII encodings of IEEE
single and double precision floating point numbers, respectively.  All
@var{int} and @var{long} values are ASCII encodings of unsigned
quantities whose ranges depend on your machine and operating system.

@menu
* Parameters File::     Defines an exponential model
* Events File::         Specifies behavior of features
* Expressions File::    Defines a probability expression
@end menu

@node Parameters File, Events File,  ,  File Formats
@section Parameters File

Each valid parameters file defines a conditional exponential model
m(y|x)
@example
        m(y|x) = r(y|x) / Z(x) 
@end example
whose numerator r(y|x) is a product of exponentials alpha_i =
exp(lambda_i) and whose denominator Z(x) is required to satisfy the
constraint that sum_y p(y|x) = 1.
@example
        r(y|x) = prod_i alpha_i^g_i(x,y) 
        Z(x) = sum_y r(y|x) 
@end example

A parameters file begins by specifying the cardinality of Y
@var{<alphabet-size>} and the total number of parameters
@var{<number-parameters>}.  Next it includes a sequence of marginal and
conditional parameters, one per line in the following format, where
@var{<number-marginal>} is the number of marginal parameters and 
@var{<number-conditional>} is the number of conditional parameters.

@example
@cartouche
 begin.parameters <alphabet-size> <number-parameters>
  begin.marginal <number-marginal>
   <marginal-parameter>
   .
   .
   .
  end.marginal
  begin.conditional <number-conditional>
   <conditional-parameter>
   .
   .
   .
  end.conditional
 end.parameters
@end cartouche
@end example

@cindex parameter, marginal
A @emph{marginal parameter} is the parameter associated with a feature
g_i(x,y) that is independent of the context x, that is, g_i(x,y) = g_i(w,y)
for all symbols y and all contexts x and w.
@cindex parameter, conditional
A @emph{conditional parameter} is the parameter associated with a
feature g_i(x,y) whose values depend on the context, that is, g_i(x,y) !=
g_i(w,y) for some symbol y and some pair of contexts x and w.
The indices assigned to marginal and conditional parameters must not
overlap.

Each parameter is a white-space delimited triple 

@example
   i:@var{int} alpha_i:@var{double} a_i:@var{double}
@end example

where i is a natural number that uniquely identifies the feature g_i,
@cindex alpha_i
alpha_i = exp(lambda_i) is the corresponding parameter value for that
feature, and a_i is the target expectation for feature g_i() with
respect to the desired joint model m(x,y).

A reasonable initial value for the alpha_i parameters is unity.  The
target expectation for a unary feature g_i() is typically the (smoothed)
empirical expectation of that feature in the training corpus.

@cindex zero index
@cindex slack feature
The feature indices need not be ordered or consecutive.  The zero index
is reserved for the distinguished slack constraint, if it is necessary.
Although the toolkit may add a parameter with index zero to a parameters
file, no user should create a parameters file containing a parameter
whose index is zero.

@node   Events File,    Expressions File,  Parameters File,   File Formats
@section Events File

An events file contains a set of distinct marginal events y from Y,
along with a nonempty set of distinct conditional events y|x from Y|X.
These events are used to calculate the expectation of each feature with
respect to the current model.  Consequently, the events file must
specify the empirical distribution on the contexts X as well as specify
the subset of the conditional event space Y|X on which each feature is
active.

@example
@cartouche
 begin.events <number-events>
  begin.marginal <number-marginal>
   <marginal-event>
   .
   .
   .
  end.marginal
  begin.conditional <number-conditional>
   <conditional-event>
   .
   .
   .
  end.conditional
 end.events
@end cartouche
@end example

@subsection Marginal Events

Each marginal event consists of three or more white-space delimited
values

@example
 y:@var{int} n(y):@var{int} i_1:@var{int} i_2:@var{int} @dots{} i_n(y):@var{int}
@end example

where y is an element of the domain Y, n(y) is the total feature
activation on y, ie., sum_i g_i(y), and the remaining values i_1, i_2,
@dots{} i_n(y) are the indices of all marginal features whose values are
nonzero on the marginal event y.  The events file must list all and only
the marginal events y that activate a marginal feature g_i(y).  In other
words, all marginal features g_i(y) must be zero for all marginal events
y not included in the events file.

@subsection Conditional Events

Each conditional event consists of four or more white-space delimited
values

@example
 x:@var{int} y:@var{int} c(x,y):@var{int} n(x,y):@var{int} i_1:@var{int} i_2:@var{int} @dots{} i_n(x,y):@var{int}
@end example

where x is an element of the domain X, 
y is an element of the domain Y, 
c(x,y) is the observed frequency of the pair <x,y>,
n(x,y) is the total feature activation on the pair <x,y>, 
@example
        n(x,y) = sum_i g_i(x,y) 
@end example
and the remaining values i_1, i_2, @dots{} i_n(x,y) are the indices of all
conditional features whose values are nonzero on the pair <x,y>.  

The only conditional events y|x included in an events file are 
(i) those whose frequency c(x,y) is nonzero and 
(ii) those whose context x occurs with nonzero frequency 
and whose conditional activation n(x,y) is nonzero.  These requirements
are both necessary and sufficient to calculate the expectations of all
features with respect to the joint distribution f(x)p(y|x) where f(x) is
the empirical distribution on X and p(y|x) the conditional distribution
given by the model.

@subsection Restrictions

@cindex events file, restrictions
No conditional or marginal event may occur more than once in an events
file.  The marginal features that are active for a conditional event y|x
should not be listed in the conditional event.  All conditional features
g_i(x,y) must be zero for all missing y|x events where the empirical
probability f(x) of x is nonzero.  All missing y|x events must have zero
frequency.  
@cindex missing events
An event is considered @emph{missing} iff it is not included
in the events file.  Consequently, each conditional event y|x in the
events file must either have positive frequency or activate a
conditional feature in a context x that has positive frequency.  These
requirements are satisfied by all maximum entropy models reported in the
statistical language modeling literature, and are necessary in order to
achieve an efficient implementation.

The events file need not include any marginal events, although it must
include at least one conditional event.  (From a modeling point of
view, however, it's essential to include marginal features so that
p(y|x) isn't uniform for novel contexts x.)  The events file must also
activate at least one feature, that is, the events file must include a
a context with nonzero frequency that activates at least one marginal
or conditional feature.

All feature indices must be strictly positive; the zero index is
reserved for the distinguished slack feature, if it is necessary.  
@cindex n-ary features
The value of an n-ary feature g_i(x,y) = k is encoded by including k-1
copies of the index i in the list.  Consequently, the entry for the y|x
event contains exactly n(x,y)+4 white-space delimited numeric values.

@node Expressions File, , Events File, File Formats
@section Expressions File

Each valid expressions file specifies a computation whose operations are
(1) to evaluate the conditional probability of an event y|x, (2) to
accumulate the results of probability computations, (3) to multiply the
results of probability computations, and (4) to concatenate the results
of probability computations.  The expressions file format is
sufficiently general to support the evaluation of complex events
including hidden variables as well as the efficient scoring of nbest
lists.

@example
@cartouche
 <expression-file> :: begin.expressions <number-expressions>
                        <expression>^+
                      end.expressions
 <expression>      :: <event-product> | <event-sum> | <event>
 <event-product>   :: begin.product <number-terms>
                       [<event> | <event-sum>]^+
                      end.product
 <event-sum>       :: begin.sum <number-terms>
                       [<event> | <event-product>]^+
                      end.sum
 <event>           :: <conditional-event>
@end cartouche
@end example

The primitive events are the leaves of the computation tree, whose
interior nodes are products and sums.  Each primitive @var{<event>} is a
conditional event, as described above, whose frequencies are required to
be one.  Recall that the set of active features includes all active
conditional features only.  The active marginal features will be
determined by the y value of the event and the marginal events specified
in the corresponding events file.  If you wish to include a marginal
event in an expressions file, you must define a distinguished empty
context to which no conditional features apply.  Unlike an events file,
a conditional event may occur more than once in an expressions file.  A
product or sum expression with zero terms will evaluate to unity or
zero, respectively.

The simplest probability computation is to compute a chain of
conditional probabilities, for which the corresponding expressions file
would be as shown.  The result of evaluating this expressions file would
be a single number, the negative natural log of the value of the given
expression according to the given model.

@example
@cartouche
 begin.expressions 1
  begin.product <number-terms>
   <conditional-event>
   .
   .
   .
  end.product
 end.expressions
@end cartouche
@end example

A more sophisticated probability computation is to compute a chain of
conditional probabilities, where each conditional probability p(y|x) is
itself the marginal of a joint probability p(y,z|x), that is, where
p(y|x) = sum_z p(<y,z>|x).  In such a situation, the conditional events
are <y,z>|x and the corresponding expressions file is a product of sums.
Again, the result of evaluating this expressions file would be a single
number, the negative natural log of the value of the given expression
according to the given model.

@example
@cartouche
 begin.expressions 1
  begin.product <number-terms>
   begin.sum <number-terms>
    <conditional-event>
    .
    .
    .
   end.sum
   .
   .
   .
  end.product
 end.expressions
@end cartouche
@end example

@node   Background,   Index,     File Formats,    Top
@chapter Background

The maximum entropy framework is a powerful method for building
statistical models.  It is expressive, allowing modelers to easily
represent their special insights into the data generating machinery.  It
is statistically efficient, because it models the intersection of
complex events without increasing the number of parameters or
fragmenting the training data.  And it provides strong models, models
that outperform their traditional variants with much less tweaking.  For
example, the maximum entropy trigram outperforms both the interpolated
trigram (Jelinek and Mercer, 1980) and the backoff trigram (Katz, 1987)
in test set perplexity as well as in speech recognizer word error rate.
For all these reasons, the maximum entropy framework has become the
framework of choice for statistical language modeling (Lau et.al, 1993;
Berger et.al, 1996; Rosenfeld 1996).

This background chapter consists of four sections.  Firstly, we review
the maximum entropy framework, and its application to statistical
language modeling.  Secondly, we consider the art of feature design for
the simplest of all powerful language models, the maximum entropy Markov
model.  Thirdly, we briefly discuss some issues that arose in the design
of the toolkit.

@menu
* Framework::                   Review of the maximum entropy framework
* Feature Design::              How to design good features
* Toolkit Design Notes::        Notes on the toolkit design
* References::                  Selected maximum entropy references
@end menu

@node Framework, Feature Design, , Background
@section Maximum Entropy Framework

The fundamental problem of statistical modeling is to induce a joint
probability model p: X,Y -> [0,1] from a finite corpus of observations
<x_1,y_1>, ..., <x_T,y_T> drawn a discrete joint domain X,Y.

In the maximum entropy approach to statistical modeling, we first define
a set G of k binary features on X,Y.
@example
        G = @{ g_i: X,Y -> @{0,1@} for i = 1...k @} 
@end example
Each binary feature g_i(x,y) partitions the joint domain X,Y into two
sets: those points <x,y> for which g_i(x,y) is active and those for
which it is not active.@footnote{The MEMT supports n-ary features, that
is, features that take on any of a finite number of nonnegative integral
values.  Here we limit our discussion to binary features in order to
simplify the presentation.}  Next, we choose a vector a = a_1 ... a_k of
target expectations for our features.  The simplest way to do this is to
choose each target expectation a_i to be the empirical expectation f[g_i],
that is, the expectation of the feature g_i(x,y) with respect to the
empirical distribution f(x,y) defined by the training corpus.
@example
        f[g_i] = sum_<x,y> f(x,y) g_i(x,y),
@end example

@menu
* Target Distribution::   The maximum entropy distribution
* Model Class::           Exponential model class
* Conditional Models::    Conditionalizing a joint model
* Computational Tricks::  Tricks to speed up computations
@end menu

@node Target Distribution, Model Class, , Framework
@subsection Maximum Entropy Distribution

Together, the feature set G and their target expectations define a class
P of all probability distributions whose feature expectations match the
target expectations,
@example
        P = @{ p: p[g_i(x,y)] = a_i for all i = 1...k @} 
@end example
where p[g_i(x,y)] is the expectation of g_i(x,y) with respect to the
distribution p(x,y).
@example
        p[g_i(x,y)] = sum_<x,y> p(x,y) g_i(x,y)
@end example
If the target expectations are the empirical expectations, then P
contains all distributions that are equivalent to the empirical
distribution defined by our training corpus, when viewed through the
eyes of our features G.  Defining the features and choosing their target
expectations is the modeler's art.

Given such a class P, we would like to find the distribution p*(x,y) in
P that maximizes the entropy H(p) with respect to all distributions in P.
@example
        H(p) = sum_<x,y> p(x,y) -log p(x,y) 
@end example
The maximum entropy distribution p*(x,y) is the one that is most
faithful to our constraints, because it makes no additional assumptions
beyond what has been specified (Jaynes 1957,1978; Csiszar, 1991).
We also need a compact way to represent this distribution, that is, we
require a model for this distribution.

@node Model Class, Conditional Models, Target Distribution, Framework
@subsection Exponential Model Class

Now consider the class R of all exponential models m(x,y) defined over
our features G
@example
        R = @{ m: m(x,y) = r(x,y) / Z @} 
@end example
whose numerators r(x,y) are a product of exponentials, alpha_i =
exp(lambda_i), and whose denominators Z are required to obtain a
probability function.
@example
        r(x,y) = prod_i alpha_i^g_i(x,y)
        Z = sum_<x,y> r(x,y) 
@end example

This model class R has as many free parameters as there are features.
Fortunately, the intersection of the class R of exponential models over
G with the class P of desired distributions is
nonempty.@footnote{Unfortunately, it is possible to define P in such a
way that the intersection of P and R is empty.  In order that the
intersection of P and R be nonempty, it suffices that for every <x,y> in
X,Y, there is at least one p in P for which p(x,y) > 0.  This condition
is easily satisfied in practice, and so we do not dwell on it here.}
Consequently, at least one of our desired distributions has a compact
representation as an exponential model with a feasible number of
parameters.  Even better, the intersection of P and R contains the
maximum entropy distribution p*(x,y).  Best of all, the intersection of
P and R is unique!  Therefore, we need only find a single exponential
model m*(x,y) in R that satisfies the linear constraints P, and we are
assured that this model is a model of the maximum entropy distribution
p*(x,y) in P.  Now if our target expectations are the empirical
expectations, then it also happens that m*(x,y) is the maximum
likelihood model in R, and so the thorny problem of finding the maximum
entropy distribution p*(x,y) in P reduces to the easy problem of finding
the maximum likelihood model m*(x,y) in R.  This is the beauty of the
maximum entropy framework (Kullback, 1959).

@node Conditional Models, Computational Tricks, Model Class, Framework
@subsection Conditional Models

A number of difficulties arise when applying these ideas to discrete
time series problems.  The first difficulty is that we must assign
probability to strings of arbitrary length, a task for which we cannot
employ joint models over finite dimensional spaces.  For discrete time
series, we require a conditional model p(y|x) instead of a joint model
p(x,y).  Our model class R becomes 
@example
        R = @{ m: m(y|x) = r(y|x) / Z(x) @}
@end example
where
@example
        r(y|x) = prod_i alpha_i^g_i(x,y)
        Z(x) = sum_z r(z|x).  
@end example
Our constraint class P remains unchanged, although we now require a
marginal distribution p(x) on the X in order to calculate the feature
expectations m[g_i(x,y)] with respect to a joint model m(x,y) =
p(x)m(y|x).  There are many reasonable choices here, but computational
efficiency will require us to use the empirical distribution f(x) on X.

@node Computational Tricks, , Conditional Models, Framework
@subsection Computational Tricks

This is the second difficulty, namely, that maximum entropy modeling for
joint spaces is computationally infeasible.  To evaluate the probability
m(x,y) of a joint event <x,y> according to our model requires us to
enumerate the entire X,Y space in order to calculate the denominator Z.
And to evaluate the expectation m[g(x,y)] of a feature g(x,y) with
respect to the joint model m(x,y) requires us to enumerate the entire
X,Y space again.

In a speech recognition application, Y is the vocabulary of all words
and X is all history equivalence classes.  The vocabulary of a typical
speech recognition system has over 20,000 words, and its history
equivalence classes X typically consist of all word bigrams, and so the
X,Y joint space contains |Y|^3 = 8x10^12 distinct events.  

The infeasibility of repeatedly enumerating this space requires us to
make two simplifications.  The first simplification allows us to
efficiently compute the expectation m[g_i] of each feature g_i(x,y),
while the second simplification allows us to efficiently compute the
denominator Z(x) of a conditional exponential model without summing over
all the symbols in Y.

@menu
* Computing m[g_i]::    Computing the feature expectations
* Computing Z(x)::      Computing the denominators
@end menu

@node Computing m[g_i], Computing Z(x), , Computational Tricks
@subsubsection Computing m[g_i]

The first simplification is to use the empirical distribution f(x) as
our marginal on X, which gives us the joint model m(x,y) = f(x)m(y|x).
The statistical consequence of this simplification is to require our
conditional model m(y|x) to match the target feature expectations on the
observed contexts only.  The computational consequence of this
simplification is quite significant, however.  Now our feature
expectations are calculated as
@example
        m[g_i(x,y)] = sum_<x,y> m(x,y) g_i(x,y) 
                    = sum_<x,y> f(x)m(y|x) g_i(x,y) 
                    = sum_x f(x) sum_y m(y|x) g_i(x,y)
@end example
The empirical distribution f(x) is zero for novel contexts, and so the
outer sum need only iterate over the observed contexts @{ x: f(x) > 0
@}.  There are at most T such contexts in a corpus of size T.  For
the interior sum we need only iterate over those symbols y for which
g_i(x,y) is active.  For all maximum entropy language models proposed in
the literature, this is bounded by a small constant, typically one.
Therefore we have reduced the entire sum_<x,y> of size |X,Y| to a sum
over the contexts observed x, which is of worst case size T.  For a
trigram model constructed from 1.4 million words of the Switchboard
corpus with a vocabulary of 22,511 words, the size |X,Y| = 22,511^3 =
11,407,339,418,831 whereas there are only 234,009 distinct contexts in
the Switchboard corpus.  Thus, this first simplification reduces the
computation of feature expectations by seven orders of magnitude.

@node Computing Z(x), , Computing m[g_i], Computational Tricks
@subsubsection Computing Z(x)

The goal of the second simplification is efficiently compute the
denominator Z(x) of the conditional exponential model m(y|x).  We
begin by partitioning the symbols Y into two sets: the set Y_x of
symbols for which some feature is active in the context x
@example
        Y_x = @{ y: g_i(x,y) != 0 for some i @} 
@end example
and its complement Y-Y_x, the symbols for which no feature is active in
the context x
@example
        Y-Y_x = @{ y: g_i(x,y) = 0 for all i @}.
@end example
This partition allows us to simplify the Z(x) summation as follows.
@example
        Z(x) = sum_@{y in Y@} r(y|x)
             = sum_@{y in Y_x@} r(y|x) + sum_@{y in Y-Y_x@} r(y|x)
             = sum_@{y in Y_x@} r(y|x) + sum_@{y in Y-Y_x@} 1
             = sum_@{y in Y_x@} r(y|x) + |Y-Y_x|
@end example
This simplification reduces the size of the Z(x) computation from |Y| to
|Y_x|.  For the Switchboard trigram model introduced above, |Y| is
22,511 whereas |Y_x| is 12,413.  Although this two-fold speedup is
significant, it is not sufficient because we must compute Z(x) for all
contexts x observed in the training corpus, for each iteration of our
parameter estimation algorithm.

Our next step is to partition the set G of features into two sets: the
marginal features G- and the conditional features G+.  A marginal
feature g_i(x,y) is a feature whose activation depends only on the
future y, completely independent of the context x.  That is, g_i(x,y) is
a marginal feature if and only if g_i(x,y) = g_i(w,y) for all w, x, and
y.  Conversely, g_i(x,y) is a conditional feature if and only if
g_i(x,y) != g_i(w,y) for some w, x, and y.  This allows us to further
partition Y_x into two sets: Y_x+ and Y_x-.  The set Y_x+ includes all
symbols for which a conditional feature g_i(x,y) is active in the
context x,
@example
        Y_x+ = @{ y: g_i(x,y) != 0 for some g_i in G+ @} 
@end example
while the set Y_x- = Y_x - Y_x+ includes all symbols for which a
marginal feature is active but no conditional feature is active in the
context x.  This further simplifies the Z(x) summation as follows.
@example
        Z(x) = sum_@{y in Y_x@} r(y|x) + |Y-Y_x|
             = sum_@{y in Y_x+@} r(y|x) + sum_@{y in Y_x-@} r(y|x) + |Y-Y_x|
             = sum_@{y in Y_x+@} r(y|x) + sum_@{y in Y_x-@} r(y) + |Y-Y_x|
             = sum_@{y in Y_x+@} r(y|x)-r(y) + sum_@{y in Y+@} r(y) + |Y-Y_x|
@end example
where r(y) is the product of the marginal parameters that apply to r(y),
@example
        r(y) = prod_@{g_i in G-@} alpha_i g_i(y) 
@end example
and sum_@{y in Y+@} r(y) is a constant independent of x.  Therefore, we
have reduced the computation of Z(x) from a sum of |Y| terms to a sum of
2|Y_x+| terms.  For the Switchboard trigram model introduced above, |Y|
is 22,511 whereas |Y_x+| is less than 204 on average, which gives us a
net 55-fold speedup.

@node Feature Design, Toolkit Design Notes, Framework, Background
@section Exercise in Feature Design

The art of maximum entropy modeling is to define an informative set of
computationally feasible features.  Each feature defines a partition of
the conditional domain Y|X.  Consequently, we would like our features to
identify the natural equivalence classes of Y|X, that is, the subsets of
Y|X for which we can gather reliable and meaningful statistics.  The
simplest features are those which identify individual observed events
y|x.  More sophisticated features might take advantage of any structure
in X or Y, for example, the fact that the contexts X might represent a
Markov equivalence relation on strings.  The highest level of
sophistication achieved by current language modeling technology is to
define features based on a domain of hidden events Z, which we first
predict as Z|X and then condition our predictions with as Y|X,Z.

We would also like our features to be computationally feasible.  The
most intensive computation required for maximum entropy modeling is to
calculate the expectations of our features with respect to the joint
model m(x,y) = m(y|x)f(x).  For each feature g_i(x,y), we must enumerate
all <x,y> for which g_i is active and m(x,y) is nonzero.  A good
conditional model m(y|x) must assign nonzero probability to all y in Y.
Since f(x) is the empirical probability, m(x,y) is nonzero for all
contexts x observed in the training data.  In many applications Y is
large and so we simply cannot afford to repeatedly iterate over Y.
Therefore, in order for the m[g_i] computation to be feasible, g_i must
be active for at most a (small) constant number of symbols y in any
observed context x.  This requirement that features be active on only a
tiny subset of the conditional domain Y|X will oblige us to choose
concrete features over abstract ones, that is, features based directly
on the observable events.

Let us now turn to the details of feature design and selection.  First,
we consider three variants of the traditional Markov model in the
maximum entropy framework: basic Markov features, overlapping Markov
features, and complemented Markov features.  The models defined by these
features are easy to understand, easy to implement, computationally
feasible, and capable of strong predictions.  Next, we discuss a class
of non-Markovian features called triggers.

@cindex notation
Before beginning, we must define our notation for strings.  Let A be a
finite alphabet of distinct symbols, |A| = k, and let x^T in A^T denote
an arbitrary string of length T over the alphabet A.  Then x_i^j denotes
the substring of x^T that begins at position i and ends at position j.
For convenience, we abbreviate the unit length substring x_i^i as x_i,
the length t prefix of x^T as x^t, and the empty string as x_i^(i-1).
Our dependent domain X is now the set of all (equivalence classes of)
histories A^* and our independent domain Y is the alphabet A.

@menu
* Markov Features::
* Overlap and Complementation::
* Trigger Features::
@end menu

@node Markov Features, Overlap and Complementation, , Feature Design
@subsection Markov Features

@cindex Markov model
@cindex Markov features, basic
The simplest example of an informative set of computational feasible
features is the class of Markov features.  Let us therefore consider how
to define a maximum entropy Markov model for the domain of strings over
an alphabet A.  Our dependent domain Y is the set of symbols A, and our
independent domain X is the set of equivalence classes on the histories
A^*.  Recall that a Markov model of order n employs the equivalence
classes X = A^n.  In such a situation, each n-th order Markov feature
g_<w^n,z>(.,.) would identify all histories A^*w^n whose suffix is w^n
for the predicted symbol z.
@example
        g_<w^n,z>(x^p,y) = 1   if x_(p-n)^p = w^n and y = z
                         = 0   otherwise
@end example

This basic Markov model both too simple and too complex.  It is too
simple because exactly one feature activates for every conditional
event.  It fails to take advantage of the central strength of the
maximum entropy framework, namely, its ability to handle overlapping
features without increasing the number of parameters or fragmenting the
training data.  This model is also too complex because there are k^(n+1)
such features -- too many parameters to collect reliable statistics for,
and too many features to train with bounded computational resources.
Even worse, most of these features will not be activated in the training
data, and if our target expectations are the empirical expectations,
then the resulting model will assign zero probability to many unseen
events.  Yet a good model must not assign zero probability to a
logically possible event.

We simplify our model by limiting our features to those that have been
observed in our training corpus.  Since our training corpus C is of
finite size T, there are at most T-n+1 n-th order Markov features to
consider.  We can further simplify our model by restricting our features
to those whose frequency c(x,y) exceeds a given threshold c_min.
@example
        G = @{ g_<w^n,z> : c(w^n,z) > c_min @}
@end example
This reduces the number of features in our model and also ensures that
all novel events will be assigned positive probability by our model.
Now our model is both statistically and computationally feasible.

Our model is still too simple, however, because at most one feature
applies to any given event y|x.  Consequently, m(y|x) will be equal to
the empirical probability f(y|x) if a feature is active, and otherwise
m(y|x) will be uniform among all conditional events z|x for which no
feature applies.  Thus, all events y|w^n in a novel context w^n will be
assigned uniform probability 1/k.  Fortunately, the Markov property
defines a rich and natural source of overlapping features, namely, all
the lower order Markov features.

@node Overlap and Complementation, Trigger Features, Markov Features, Feature Design
@subsection Overlap and Complementation

@cindex interpolated Markov model
@cindex Markov features, overlapping
The simplest powerful n-th order Markov model in the maximum entropy
framework includes all features g_<w^i,z>(.,.), for i = 0 ... n, whose
observed frequency c(w^i,z) exceeds a given threshold c_min.
@example
        G = @{ g_<w^i,z> :  0 <= i <= n,  c(w^i,z) > c_min @}
@end example
There are a feasible number of such features, and the number of active
features will vary from 0 to n+1.  Best of all, an event y|w^n with a
novel context w^n will only be assigned uniform probability if all
suffixes of w^n are novel, and the unigram distribution in the training
data is uniform.  This is the maximum entropy implementation of the
interpolated Markov model (Jelinek and Mercer, 1980).

@cindex Markov features, complemented 
An alternate interpretation for our Markov features is to say that a
lower order feature applies only if no higher order feature applies.
Under this @emph{complemented} interpretation, at most one Markov
feature applies to any given conditional event.  This complemented
Markov model is computationally efficient, because at most one feature
applies to every conditional event.  (Indeed, we can directly calculate
the parameter values for such a model without using an iterative
estimation algorithm.)  And it does not suffer the statistical problems
of the basic Markov model.  Every conditional event y|x that activates
at least one feature in the overlapping Markov model will activate
exactly one feature in the complemented Markov model.

@cindex Markov features, heterogeneous
A final refinement is to use both overlapping and complemented features
(Ristad and Thomas, 1995).  The complemented features provide sharper
estimates for novel contexts while the overlapping features provide
smoothing.  In such a heterogeneous model, a lower order overlapping
Markov feature includes all events for which we have a more specific
Markov feature as well as many events for which we lack a more specific
feature.  Adding the complemented features to an overlapping Markov
model provides sharper estimates for those events which lack a more
specific feature.  Or, equivalently, adding the overlapping features to
a complemented Markov model provides smoother estimates for all events.

@node Trigger Features, , Overlap and Complementation, Feature Design
@subsection Trigger features

Up to now, our modeling hasn't taken advantage of the maximum entropy
framework.  There is nothing about our Markov models that can't be done
equally effectively (and more efficiently) using traditional techniques.
So as the final section of our exercise in feature design, let us
introduce a simple class of non-Markovian features called triggers.  A
trigger is function of the entire history, not just the last n symbols.
In the trigger model (Lau et.al. 1993; Rosenfeld 1996), a trigger
function d_w(x^t) is activated if and only if the word w occurs
somewhere in the history x^t.  When the trigger words are chosen
appropriately, the trigger functions model the fact that discourses have
topics, that certain words are strong indicators of the underlying
topic, and that the topic influences the language user's choice of
words.

Recall that a Markov model of order n maps each history x^t into its
equivalence class x_(t+1-n)^t.  In an alphabet of size k, there are k^n
such history equivalence classes along with k symbols to predict, and
therefore a basic Markov model of order n has k^(n+1) parameters.  A
trigger model of order (n,l) consists of an order n Markov model along
with l trigger functions.  It maps each history x^t into its equivalence
class
@example
        <x_(t+1-n)^t, d_1(x^t), ..., d_l(x^t)>
@end example
where each trigger d_i:A^* -> @{0,1@} is a binary function on the set
A^* of possible histories.  Therefore the trigger model defines k^n 2^l
equivalence classes of histories.  Implementing the trigger model using
traditional techniques would require k^(n+1) 2^l parameters, ie., far
too many parameters to estimate reliably.  Indeed, nearly all of our
contexts will be unique.  Fortunately, the maximum entropy framework
suggests an elegant solution to this problem.

@cindex trigger feature
The maximum entropy implementation of the trigger model consists of two
classes of features: Markov features (as described above) and trigger
features.  The trigger features are defined most generally as
@example
        g_<i,z>(x^p,y) = 1   if d_i(x^p) = 1 and y = z
                       = 0   otherwise
@end example
where for the Lau et.al. (1993) and Rosenfeld (1996) word triggers,
d_w(x^p) activates iff x^p is in A^*wA^*.  This model uses the same
system of k^n 2^l equivalence classes but contains only (k^n + l)k
parameters.  This is the statistical efficiency of the maximum entropy
framework.

@node Toolkit Design Notes, References, Feature Design, Background
@section Toolkit Design Notes

@cindex file-based interface
The toolkit was designed to be used in a collaborative research setting.
A file-based interface is ideal in such a setting because it does not
limit the user's software environment -- their choice of programming
style, language, compiler, operating system, or machine -- in any way.
Everyone is free to make their own choices without retarding the group
effort.  A file-based interface provides an automatic "paper trail" for
debugging and check-pointing, and a monotonic path to a successful
implementation.  Bugs can always be fixed manually, by hand-editing the
files.  Unfortunately, a file-based interface is likely to be too
inefficient in both time and space to be used in an application.

@cindex c(x,y)
@cindex c(x)
The Events File requires the user to specify the frequency c(x,y) of
every conditional event y|x in the training corpus.  This information is
used to compute the empirical distribution f(x) = sum_y c(x,y) / T over
the contexts X.  It is also necessary to compute the total codelength
L(C|m) of the corpus C given the model m.  A more spartan design would
only require the user to provide the frequency c(x) of every context;
this suffices to perform estimation.  However it is not possible to
compute L(C|m) without c(x,y), and L(C|m) is an important convergence
measure for @code{me.estimate} (see above).  Therefore we chose to
require the full c(x,y) instead of only requiring c(x).

@cindex minimum divergence
Our initial design goal was to implement the minimum divergence
framework.  The minimum divergence framework is a generalization of the
maximum entropy framework, where we introduce a reference distribution
q(x,y) and then search for the distribution p_q^*(x,y) that satisfies a
set of linear constraints P = @{ p: p[g_i] = a_i @} and whose divergence
D(p||q) with respect to the reference distribution q(x,y) is minimal
among all distributions in P.  When the reference distribution is
uniform, then the minimum divergence distribution is identical to the
maximum entropy distribution.  The fundamental software difficulty posed
by the minimum divergence framework is to design a clean and efficient
file-based interface to the reference distribution.  The simplest
interface would require the user to supply a reference probability
q(y|x) for @emph{every} conditional event y|x, including those that
never occurred and those for which no feature was active.  Clearly this
is not feasible, either in time or space, and so an effective design of
a minimum divergence toolkit must strike some compromise between
generality and feasibility.  Rather than strike any compromises at this
point, we decided to limit our support to the maximum entropy framework.

@cindex Events file design flaw
The central flaw in the MEMT design lies in the events file format,
which results in impractically large files.  An events file completely
specifies the behavior of our features on the training corpus.  As such,
it must not only summarize the training corpus, but it must also
enumerate all conditional events y|x for which some conditional feature
is active and the context x was observed in the training corpus.  This
has the advantage of simplicity, but the disadvantage of infeasibility
because such an events file may contain a great many unobserved events.
In the worst case, the events file is size O(T V) for a training corpus
containing T observations and a vocabulary of size V.  In many cases,
this is infeasibly large.  For example, a large word trigram model
induced from 1.4 million words of Switchboard text over a vocabulary of
22,511 words contains 12,413 unigram features, 80,643 bigram features
and 120,116 trigram features.  The corresponding events file is 1gb
because every observed context x generates on average 204.3 events, of
which only 2.0 are actually observed in the training corpus.  The
remaining 202.3 events per observed context do not occur in the training
corpus.  Thus, the events file is two orders of magnitude larger than
the training corpus.  The next major release of the MEMT will include a
redesign of the events file format to redress this flaw.

@node References, , Toolkit Design Notes, Background
@section References

@enumerate
@item
A. Berger, S. Della Pietra, and V. Della Pietra.
A maximum entropy approach to natural language processing.
@emph{Computational Linguistics}, 22(1):39-71, 1996.

@item
I. Csiszar. 
Why least squares and maximum entropy?  
An axiomatic approach to inference.
@emph{Annals of Statistics}, 19(4):2032-2067, 1991.

@item
I. Csiszar and G. Longo.
Information geometry and alternating minimization procedures.
@emph{Statistics and Decisions}, Supplement Issue 1:205-237, 1984.

@item
I. Csiszar.
A geometric interpretation of Darroch and Ratcliff's generalized iterative scaling.
@emph{The Annals of Statistics}, 17(3):1409-1413, 1989.

@item
J.N. Darroch and D. Ratcliff.
Generalized iterative scaling for log-linear models.
@emph{The Annals of Mathematical Statistics}, 43(5):1470-1480, 1972.

@item
S. Della Pietra, V. Della Pietra, and J. Lafferty.
Inducing features of random fields.
Technical Report CMU-CS-95-144, CMU, Pittsburgh, May 1995.

@item
J.J. Godfrey, E.C. Holliman, and J. McDaniel.
Switchboard: telephone speech corpus for research and development.
In @emph{Proc. IEEE ICASSP}, pages 517-520, Detroit, 1995.

@item
E. T. Jaynes.
Information theory and statistical mechanics.
@emph{Phys. Rev.} 106: 620-630; 108: 171-182, 1957.

@item
E. T. Jaynes.
Where do we stand on maximum entropy?
In R. D. Levine and M. Tribus, editors, 
@emph{The Maximum Entropy Formalism}. MIT Press, Cambridge, MA, 1978.

@item
F. Jelinek and R. L. Mercer.
Interpolated estimation of Markov source parameters from sparse data.
In Edzard S. Gelsema and Laveen N. Kanal, editors, 
@emph{Pattern Recognition in Practice}, 
pages 381-397, Amsterdam, May 21-23 1980. North Holland.

@item
S. Katz.
Estimation of probabilities from sparse data for the language model
component of a speech recognizer.
@emph{IEEE Trans. ASSP}, 35:400-401, 1987.

@item
S. Kullback.
@emph{Information Theory and Statistics}.
Wiley, New York, 1959.

@item
R. Lau, R. Rosenfeld, and S. Roukos.
Trigger-based language models: a maximum entropy approach.
In @emph{Proc. ICASSP-93}, pages 45-48, 1993. vol.II.

@item
R. Rosenfeld.
A maximum entropy approach to adaptive statistical language modeling.
@emph{Computer, Speech, and Language}, 10:187-228, 1996.

@item
E. S. Ristad and R. G. Thomas.
Nonmonotonic extension models.  Research Report CS-TR-486-95,
Department of Computer Science, Princeton University, February 1995

@item
E. S. Ristad and P. N. Yianilos.
The library of practical abstractions, release 1.2.
ftp://ftp.cs.princeton.edu/pub/packages/libpa. 
December 1996.

@end enumerate

@node Index, , Background, Top
@unnumbered Index
@printindex cp

@contents
@bye